\pdfoutput=1

\documentclass[11pt]{article}

\usepackage[]{acl}

\usepackage{times}
\usepackage{latexsym}

\usepackage{soul}
\usepackage{url}
\usepackage{graphicx}
\usepackage{amsmath}
\usepackage{amsthm}
\usepackage{amssymb}
\usepackage{bm}
\usepackage{booktabs}
\usepackage{algorithm}
\usepackage{algorithmic}
\usepackage{subfigure}
\usepackage{multirow}
\usepackage{enumitem}

\usepackage[T1]{fontenc}

\usepackage[utf8]{inputenc}

\usepackage{microtype}

\title{Cross-Lingual Ability of Multilingual Masked Language Models: \\
A Study of Language Structure}

\author{Yuan Chai$^{1,}$\thanks{\quad Work done during internship at Microsoft Research Asia.}, Yaobo Liang$^{2}$, Nan Duan$^{2}$ \\
  $^1$ Beihang University \\
  $^2$ Microsoft Research Asia  \\
\tt $^1$chaiyuan@buaa.edu.cn \\
\tt $^{2}$\{yalia, nanduan\}@microsoft.com
}

\begin{document}
\maketitle
\begin{abstract}
Multilingual pre-trained language models, such as mBERT and XLM-R, have shown impressive cross-lingual ability. Surprisingly, both of them use multilingual masked language model (MLM) without any cross-lingual supervision or aligned data. Despite the encouraging results, we still lack a clear understanding of why cross-lingual ability could emerge from multilingual MLM. In our work, we argue that cross-language ability comes from the commonality between languages. Specifically, we study three language properties: constituent order, composition and word co-occurrence. First, we create an artificial language by modifying property in source language. Then we study the contribution of modified property through the change of cross-language transfer results on target language. We conduct experiments on six languages and two cross-lingual NLP tasks (textual entailment, sentence retrieval). Our main conclusion is that the contribution of constituent order and word co-occurrence is limited, while the composition is more crucial to the success of cross-linguistic transfer.
\end{abstract}
\section{Introduction}

Zero-Shot Cross-Lingual transfer aims to build models for the target language by reusing knowledge learned from the source language. In this way, models can be efficiently implemented in multilingual as well as low-resource language scenarios. Traditionally, it is solved by a two-step pipeline \cite{ruder2019survey}: a shared multilingual textual representation is first built and then supervised data from the source language is used on the top of it to train task-specific models. With the recent emergence of multilingual language models, the standard paradigm in this field has shifted to the pre-trained fine-tuning paradigm. Multilingual pre-trained language models, such as mBERT \cite{devlin-etal-2019-bert} and XLM-R \cite{conneau2020unsupervised}, have proven effective for cross-lingual transfer with better results on a large number of downstream tasks and languages \cite{pires2019multilingual, conneau2020unsupervised}. \par
\begin{figure}[t]
    \centering
    \includegraphics[width=0.85\columnwidth,]{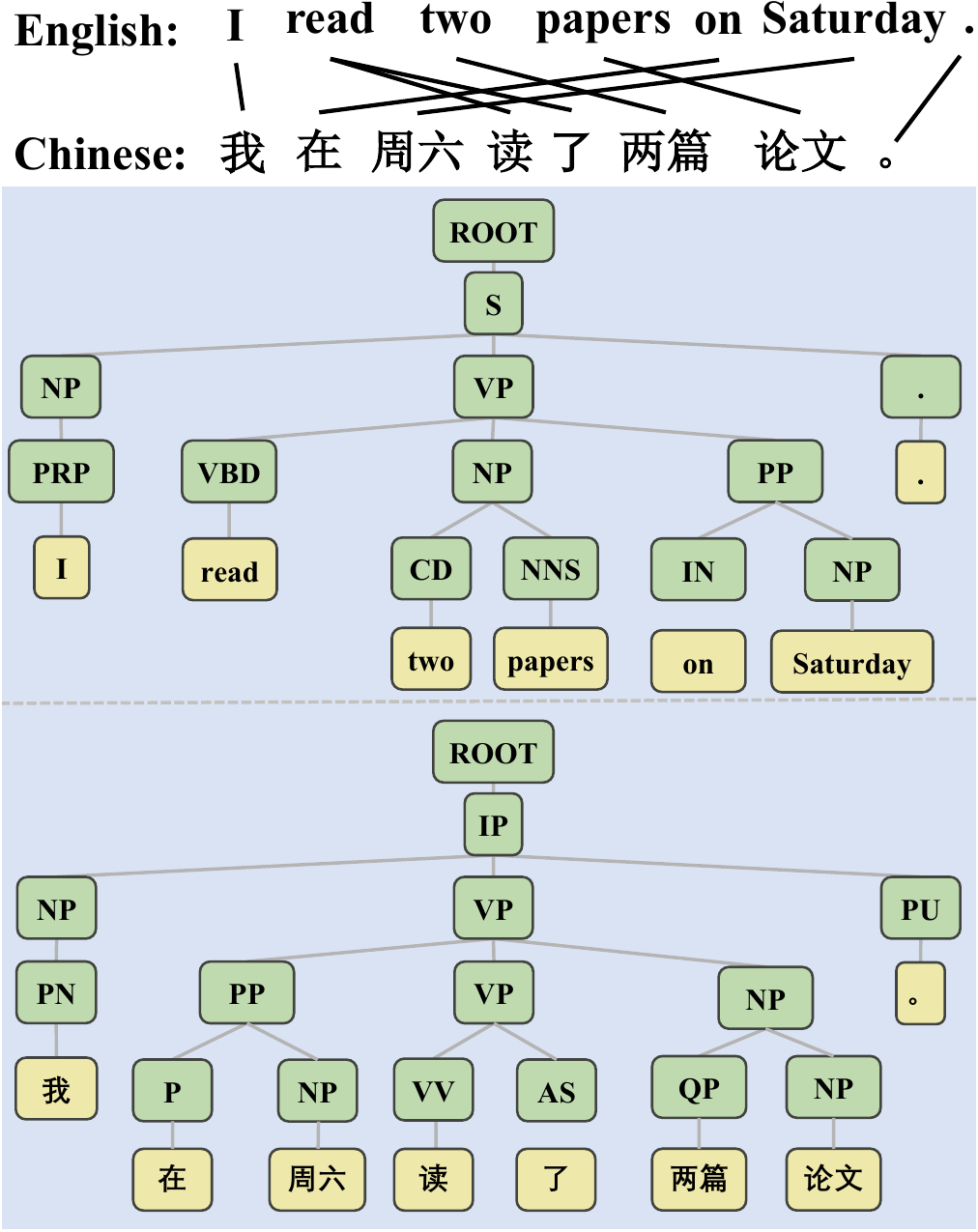}
    \caption{Example of two sentences in different languages, which are different in constituent order, but are very similar in constituent tree. Note that we simplify the constituent tree for better understanding.}
    \label{fig:intro-1}
\end{figure}
The most surprising part is that mBERT and XLM-R are both trained without using any parallel corpus. Previous work \cite{pires2019multilingual, wu-dredze-2019-beto} attribute this success to the shared anchor words. But recent work \cite{conneau2020emerging, Karthikeyan2020cross} shows that cross-lingual transfer still could emerge even corpus of languages are from different domain, or don't share any common words. For cross-lingual model, sharing transformer encoder weight is critical, while whether having language specific word embedding or language identity marker is not important. This makes us more curious about what kind of common property of languages could make cross-lingual transfer successful.

In our work, we study three language structure properties: 1) Constituent order. Specifically, we study three common constituent orders: order of verb and object, adposition and noun phrase, adjective and noun. 2) Composition. Composition means that we could combine two or several simple meanings and build a new more complicated meaning. For example, two words could form a phrase, and more composition could form a sentence recursively. 3) Word co-occurrence. We take the bag of words assumption and study the word co-occurrence in a sentence.

We use Figure \ref{fig:intro-1} to better show the composition similarity between two sentences. Although the English sentence and Chinese sentence have different word order, but they are both first divided into a noun phrase and a verb phrase, and the verb phrase is then divided into three parts that are identical in meaning but inconsistent in order.

To better analyze the contribution of these three properties, we use the control variable method. Based on a successful transfer between the source and target languages, we change or remove only one structure property in the source language. We measure the importance of this property by testing the change in performance of the cross-language transfer from the modified source language to the target language. The results show that the effect of constituent order and word co-occurrence is small, while composition has a greater effect. \par

The main contributions are summarized as follows: (1) We analyze the source of cross-linguistic ability from shared properties in language structure and propose three candidate answers. (2) We used the control variable method, modifying only the target property in the corpus and keeping the other settings identical, thus better quantifying the contribution of the studied properties. (3) Our experiments clearly show that constituent order and word co-occurrence make very limited contributions to cross-lingual ability, while composition is the key to cross-lingual transfer.

\section{Study Design}

In this section, we introduce the design of our study, including the three language structure properties, and the overall setup. We also detail the pre-training and fine-tuning settings for better reproduction.
\subsection{Dissecting Language Structure}
\paragraph{Constituent Order} In constituent tree, the constituents in a grammar rule are often ordered. For example in English, "{\ttfamily S->NP VP}" means that we should put the noun phrase at beginning of sentence and put the verb phrase after it. There are many linguistic studies to summarize and compare the constituent order of different languages, such as WALS \cite{wals}. We mainly study three WALS features, 83A (Order of Object and Verb), 85A (Order of Adposition and Noun), and 87A (Order of Adjective and Noun).
\paragraph{Composition} Composition means to combine two or several meanings and build a new more complicated meaning. As shown in Figure \ref{fig:intro-1}, "two" and "papers" could form a new meaning "two papers". And we could further combine it with "read" to form "read two papers". To better dissect the language structure, the composition in our study doesn't have order. By recursively combining meanings, we could express infinite meanings with finite words. The combination process forms an unordered tree. 
\paragraph{Word Co-Occurrence} In our paper, we study the word co-occurrence at the sentence level. Some words often co-occur in a context window or a sentence. As shown in Figure \ref{fig:intro-1}, the word co-occurrence of sentences with same meaning may also be similar in different languages, which may be a source of cross-lingual ability.

The natural language sentences of most languages are composed of a list of ordered words. But different languages may have different word order. We argue that most research about word order, for example research in WALS, are studying constituent order. The term "word order" hypothesis that any word could have any neighboring word. But "constituent order" hypothesis that some words should always group together and form a constituent and the order between groups are the object of study. The words from two constituents are unlikely to be neighboring words. Based on this, we dissect the "word order" into two concepts "constituent order" and "composition". "composition" is the rules to group words to phrase, clause and sentence. If we remove all the word order information, we will only have a set of unordered words and we name this feature as word co-occurrence. Bag-of-Word assumption only takes word co-occurrence information and has achieved great success in topic modeling and word embedding. We also hope to study its influence to cross-lingual ability of MLM.
\subsection{Overall Setup}
\paragraph{Bilingual Pre-training} Following previous studies \cite{conneau2020emerging,Karthikeyan2020cross}, our experiments were done on the corpus of only two languages, source (English) and target (multiple languages). By involving only one pair of languages, we can ensure that the performance of a given target language is only affected by the source language, without worrying about interference from a third language. In our work, we select English as source language because it has best constituent parser and most of cross-lingual benchmarks only have English training data. \par
\paragraph{Only Modify Source Language} We believe the source of cross-lingual ability is the commonality between languages, and it can be destroyed by modifying either language in the pair. We decide to only modify source language and leave target language unmodified. This makes results on target language comparable to each other and ensures that changes in the results only come from changes of language property rather than modifications in the target language. By keeping the other settings the same and modifying only the source language, we exclude the interference of extraneous factors. This setup follows the control variable method and allows a more precise quantification. \par
\paragraph{Consistent in Pre-training and Evaluation} We study the different commonality by creating a new language. So we make modifications both in pre-training and downstream evaluation. This consistency could help to generalize our conclusions to new languages beyond the 100 human natural languages. For example, the new languages could be other modalities like image, audio and video. Or programming languages like Python, Java and Lisp. We may meet extra-terrestrial someday and could access their unlabeled textual corpus. We still hope cross-lingual research could help us to understand their languages.

\subsection{Multilingual Masked Language Model}

Our multilingual masked language model pre-training follows the standard setup such as mBERT, XLM-R.  Specifically, we mask 15\% of the input tokens, of which 80\% are replaced with mask tokens, 10\% keep the original words, and 10\% are randomly replaced with sampled words from the multilingual vocabulary. The training objective is to recover the masked tokens. 
We use the entire Wikipedia for each language as pre-training data and the model parameters are shared across languages. Unlike standard multilingual pre-trained models, the vocabulary in our experiments is not shared across languages. To remove confounding factors, our vocabulary is learned individually on each language using BPE \cite{sennrich2016neural}, as  \cite{conneau2020emerging,Karthikeyan2020cross} have demonstrated that sharing vocabulary has a very limited effect on cross-lingual transfer.  Note the softmax prediction layer shared across languages is still preserved. 

\paragraph{Implementation Details} We use base size model in each experiment, which is a Transformer \cite{vaswani2017attention} with 12 layers, 12 heads, and GELU activation functions. The vocabulary size is 32k for each language, the embedding dimension is 768, the hidden dimension of the feed-forward layer is 3072, and the dropout rate is 0.1. We use the Adam optimizer and the polynomial decay learning rate scheduler with ${\rm 3\times10^{-4}}$ learning rate and 10k linear warm-up steps during training. We train each model with 8 NVIDIA 32GB V100 GPUs and use total batch size 2048 with gradient accumulation strategy. We stop pre-training at 160k steps and evaluate the pre-trained model on downstream tasks every 8k steps and report the best result.

\subsection{Downstream Cross-lingual Evaluation}
We consider Cross-lingual Natural Language Inference (XNLI) dataset \cite{conneau2018xnli} and Tatoeba dataset \cite{artetxe2019massively} in XTREME benchmark \cite{hu2020xtreme} to evaluate performance.

\paragraph{XNLI} is a standard cross-lingual textual entailment dataset, which asks whether a premise sentence entails, contradicts, or is neutral toward a hypothesis sentence in the same language. We use the \textbf{zero-shot cross-lingual transfer} setting, where we first fine-tune the pre-trained model with source (English) language and then directly test the model with target language. XNLI is a three-category classification task which uses accuracy as its metric. The three categories in the test set are uniformly distributed, so the score of random guesses is 33.33\%.

\paragraph{Tatoeba} is a cross-lingual sentence retrieval dataset which consists of up to 1,000 English-aligned sentence pairs covering 122 languages. Tatoeba uses the source to target Top-1 accuracy as its metric. Note that Tatoeba only has test set so we use the pre-trained model directly without fine-tuning.

\paragraph{Evaluation Details} For XNLI, the task-specific layer is a two layer linear mapping with tanh function between them, which takes the {\ttfamily[cls]} token as input. We use the Adam optimizer and linear decay learning rate scheduler with ${\rm 7\times10^{-6}}$ learning rate and 12.5k linear warmup steps during fine-tuning. We fine-tune each model with batch size 32 for 10 epochs and evaluate on the English dev set every 3k steps to select the best model. We report the result on average of four random seed.
For Tatoeba, we use the average pooling subword representation (excluding special token) of sentences at the 8-th layer as sentence representations following XTREME settings ~\cite{hu2020xtreme}. Evaluation is done by finding the nearest neighbor for each sentence in the other language according to cosine similarity. 

\begin{figure*}[ht]
    \centering
    \includegraphics[width=\textwidth,]{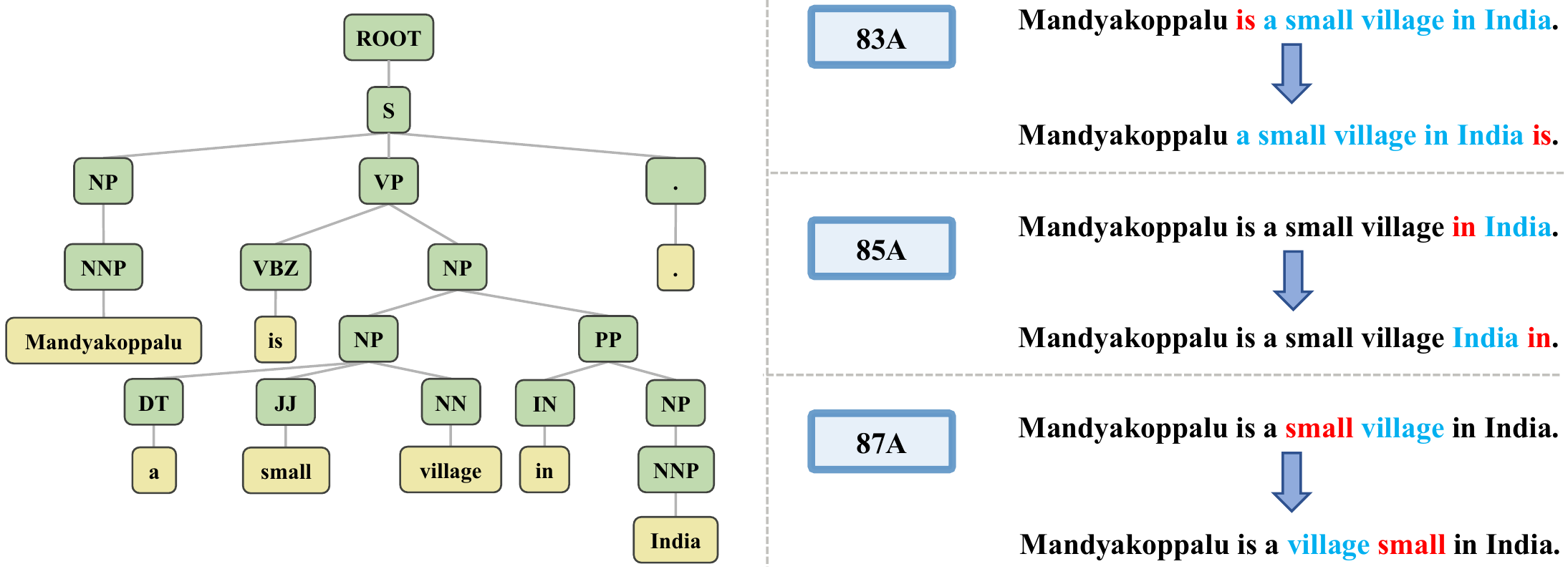}
    \caption{Example of constituent tree and constituent order modification based on WALS features. 83A, 85A and 87A represent order of object and verb, order of adposition and noun phrase and order of adjective and noun.}
    \label{fig:example1}
\end{figure*}

\section {Constituent Order}
Previous work\cite{pires2019multilingual} has argued that the cross-lingual transfer performance between languages with same constituent order is 10\%-20\% better than languages with different constituent order. So we further conduct control variate experiments to study the influence of constituent order. First we introduce the constituent order we studied and experiment setup. Then we analyze the effects of constituent order through the results. Our main conclusion is that the contribution of constituent order is about 1\%.

\subsection{Constituent Order Modification}
Following \cite{naseem2012selective,pires2019multilingual}, we use a subset of order-related features from WALS to study constituent order. Specifically, we examine:
\begin{itemize}[leftmargin=*]
    \item \textbf{Order of Object and Verb.} Corresponding to 83A in WALS and grammar "{\ttfamily VP->VB NP}" in the constituent tree. Two orders are defined in WALS, OV for Object-Verb order and VO for Verb-Object order. English is an OV language and we change it to VO by changing the grammar to "{\ttfamily VP->NP VB}". Note that we consider all tags starting with {\ttfamily VB} ({\ttfamily VBZ}, {\ttfamily VBD}) as {\ttfamily VB}.
    \item \textbf{Order of Adposition and Noun Phrase.} Corresponding to 85A in WALS and grammar "{\ttfamily PP->IN NP}" in the constituent tree. Two orders are defined in WALS, Prepositions (Pre) for Preposition-Noun Phrase order and Postpositions (Post) for Noun Phrase-Postposition order. English is a Prepositions language and we change it to Postpositions by changing the grammar to "{\ttfamily PP->NP IN}".
    \item \textbf{Order of Adjective and Noun.} Corresponding to 87A in WALS and grammar "{\ttfamily NP->JJ NN}" in the constituent tree. Two orders are defined in WALS, AN for Adjective-Noun order and NA for Noun-Adjective order. English is a AN language and we change it to NA by changing the grammar to "{\ttfamily NP->NN JJ}".
\end{itemize}

Specifically, we use the Constituency Parsing tool in Stanford's CoreNLP \cite{manning2014stanford} to obtain the constituent trees. For each order-related feature, we filter out the parent node and children nodes satisfy the feature's grammar. For example, the grammar for order of object and verb is "{\ttfamily VP->VB NP}". We filter out the parent node, whose constituent label is VP, with and only with two children nodes, whose constituent labels are VB and NP respectively. Then we will change the order of the two children nodes. After we recursively check and modify all tree nodes, we in-order traverse the tree and get the sentence with constituent order modified. In Figure \ref{fig:example1}, we show examples of modifying constituent order. 

We select Spanish, Russian, Hindi, Turkish, Thai, and Vietnamese as target languages, considering the variance of script, typological features and pre-training resources. 

Unlike the analysis of correlations between constituent order and results of target languages \cite{pires2019multilingual}, we follow the principle of control variables and modify the constituent order directly in the corpus. In this way, we can ensure that the differences in results come from constituent order modifications only.

\begin{table*}[!ht]
\centering
\resizebox{0.7\textwidth}{!}{\scriptsize
\begin{tabular}{l|cc|cc|cc}
\hline
\multirow{2}{*}{source} & \multicolumn{2}{c|}{ru (VO, Pre, AN)} & \multicolumn{2}{c|}{hi (OV, Post, AN)} & \multicolumn{2}{c}{tr (OV, Post, AN)} \\ \cline{2-7} 
                             & source     & target     & source     & target     & source     & target    \\ \hline
en                           & 83.93      & 73.90      & 83.83      & 69.46      & 83.92      & 72.48     \\
en-OV                        & 83.81      & 74.03      & 83.24      & 69.66      & 83.81      & 73.27     \\
en-Post                      & 83.56      & 74.09      & 83.68      & 70.20      & 83.93      & 72.84     \\
en-NA                        & 83.64      & 73.85      & 83.54      & 69.56      & 84.07      & 72.88     \\ \hline
\multirow{2}{*}{source} & \multicolumn{2}{c|}{es (VO, Pre, NA)} & \multicolumn{2}{c|}{th (VO, Pre, NA)} & \multicolumn{2}{c}{vi (VO, Pre, NA)} \\ \cline{2-7} 
                             & source     & target     & source     & target     & source     & target    \\ \hline
en                           & 83.97      & 78.11      & 83.29      & 70.21      & 83.87      & 74.55     \\
en-OV                        & 83.81      & 78.76      & 83.20      & 70.25      & 82.76      & 74.78     \\
en-Post                      & 83.55      & 78.22      & 83.80      & 70.31      & 83.64      & 74.34     \\
en-NA                        & 84.01      & 78.18      & 84.01      & 71.34      & 83.94      & 74.86     \\ \hline
\end{tabular}}
\caption{Comparison of XNLI results before and after modification of constituent order. "en" represents the unmodified English, en-OV represents the modification of English from Verb-Object order to Object-Verb, and the same for the others. We marked its original constituent order after each target language.}
\label{tab:xnli-1}
\end{table*}

\begin{table}[h]
\centering
\resizebox{0.7\columnwidth}{!}{\scriptsize
\begin{tabular}{l|ccc}
\hline
source          & ru   & hi   & tr  \\ \hline
en              & 78.1 & 60.6 & 58.9 \\
en-OV           & 73.6 & 66.9 & 64.5 \\
en-Post         & 70.1 & 65.0 & 60.4 \\
en-NA           & 77.7 & 59.8 & 59.9 \\ \hline
source          & es   & th   & vi   \\ \hline
en              & 82.2 & 59.9 & 76.0 \\
en-OV           & 78.1 & 54.6 & 68.7 \\
en-Post         & 80.4 & 51.8 & 72.7 \\
en-NA           & 76.8 & 52.6 & 78.5 \\ \hline
\end{tabular}}
\caption{Comparison of Tatoeba results before and after modification of constituent order.}
\label{tab:Tatoeba-1}
\end{table}

\subsection{Effect of Constituent Order}
Table \ref{tab:xnli-1} and Table \ref{tab:Tatoeba-1} show the results on XNLI and Tatoeba. With the results of modifying three features and test on six different target languages, we can draw the following three conclusions: \par
\textbf{Modifying constituent order barely affects source language.} As shown in \ref{tab:xnli-1}, modifying constituent order in the source language barely change its XNLI results (basically 0.3\%). This means that our modifications do not affect the overall meaning of the language. The modified language is still a reasonable language for both humans and models. \par 
\textbf{Changing source language's constituent order to same as target language could improve cross-lingual transfer.} In Tables \ref{tab:xnli-1} and \ref{tab:Tatoeba-1}, we find that modifying constituent order achieves consistent gains on most low-resource languages. For example, modifying 83A in Turkish achieves gains 0.79\% on XNLI and 5.6\% on Tatoeba , 85A in Hindi gains 0.74\% on XNLI and 4.4\% on Tatoeba , 87A in Vietnamese gains 0.31\% on XNLI and 2.5\% on Tatoeba . However, this pattern is not very stable in high-resource languages. For example, 87A in Spanish gains only 0.08\% on XNLI and decreases by 5.4\% on Tatoeba instead.

\textbf{The overall effect of modifying constituent order to cross-lingual transfer is limited.} No matter what modification is made, the results on six different target languages showed very limited changes (basically within 1\% on XNLI and 8\% on Tatoeba). This further suggests that constituent order has limited effect on cross-lingual transfer. In other words, constituent order is not the key component of language structure.
As for the magnitude of the variation, it is slightly higher on Tatoeba than on XNLI. We believe there are two main reasons. First, the average sentence length of Tatoeba is lower than XNLI, so the effect of modification will be magnified. Second, Tateoba doesn't have training data and the zero-shot evaluations are highly unstable. For example, \cite{phang2020english} achieved more than 20\% gains by fine-tuning the model on XNLI at first. \par

\subsection{Comparison to Previous Work}
The "conflict" conclusion between our work and \citealt{pires2019multilingual} is because of the difference of experiment design. In the experiment about object and verb order, \citealt{pires2019multilingual} change the source language to totally different language, and test on target languages. For example train on English (VO) or Hindi (OV), and test on French (VO). And they found the transfer from VO to VO is much better than transfer from OV to VO. While our experiments use modified English. We argue that the verb and object order isn't the only difference of source language in their experiment. For example, most of Europe languages are VO and most of Central Asia languages are OV. Languages in the same region are more similar than languages in different region. Our work conduct control variate experiments and could analysis the importance of constituent order better.

\section{Composition: The Key to Zero-Shot Cross-Lingual Transfer}
In this section, we study the contribution of constituent order, composition and word co-occurrence respectively.  We first present how to completely remove constituent order and composition step-by-step from the corpus, and then analyze the results. Subsequently, by controlling the rate of composition retention, we further quantified its contribution to cross-lingual transfer.

\subsection{Language Ablation of Removing Constituent Order and Composition}
\begin{figure*}[ht]
    \centering
    \includegraphics[width=\textwidth,]{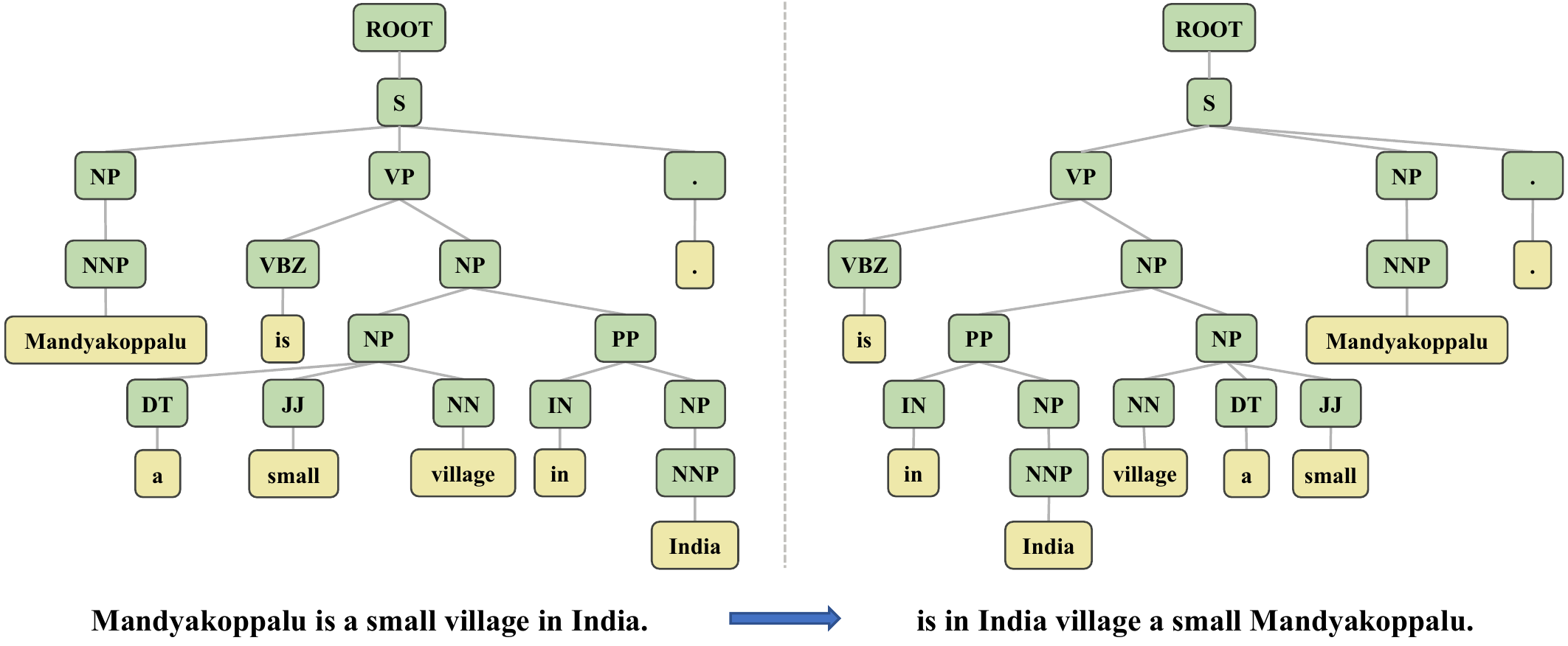}
    \caption{Example of constituent shuffle by disordering children of each intermediate node in constituent tree. }
    \label{fig:example2}
\end{figure*}

First, we introduce several experiments settings:

\textbf{Constituent Shuffle: Removing Constituent Order.}
When removing the constituent order, we should be careful to keep the composition untouched. As shown in Figure \ref{fig:example2}, we shuffle the children nodes of same intermediate node in the constituent tree, while preserving the parent-children relation between nodes unchanged. By comparing its results with the baseline, we can quantify the contribution of constituent order. 

\textbf{Word Shuffle: Removing Constituent Order and Composition.}
To further remove the composition, we randomly shuffle the words in sentence. This "Word Shuffle" operation will remove constituent order and composition together. By comparing it with the results of "constituent shuffle", we can quantify the contribution of composition. 

\textbf{Baselines Without Pre-training: Removing Constituent Order, Composition and Word Co-occurence.}
We also provide a "Without Pre-training" baseline in XNLI and "Word Embedding Average" baseline in Tatoeba to quantify the contribution of word co-occurrence by comparing with "Word Shuffle". On XNLI, "Without Pre-training" represents a Transformer model with same structure as pre-trained model but with random initialized weights. Then we fine-tune it with source language and test on target languages. Because Tatoeba doesn't have any training data, we use the average of word embedding as a baseline. The word embedding is extracted from the embedding layer of "word shuffle" setting. The performance of word embedding average baseline still credits to word-occurrence but not to pre-training.  \par

Second, to quantify the modification degree, we define two metrics. Inversion Ratio is the number of inverse pairs in the modified sentence, which normalized by the number of total word pairs in the sentence. Word Move Distance is the average distance of each word moved in sentence, which normalized by length of each sentence. As shown in Table \ref{tab:shuffle-metric}, the sentences after constituent shuffle and word shuffle are almost identical in two metrics and both much higher than sentences modifying local constituent order. This shows that constituent shuffle also makes lots of word order modifications and has high randomness. \par

\begin{table}[h]
\centering
\resizebox{0.9\columnwidth}{!}{\scriptsize
\begin{tabular}{l|c|c}
\hline
source type         & IR (\%) &  WMD (\%) \\ \hline
en-OV               & 3.04                 & 2.71                    \\
en-Post             & 4.26                 & 3.66                    \\
en-AN               & 0.49                 & 0.35                    \\
Constituent Shuffle & 50.44                & 31.1                    \\
Word Shuffle        & 51.89                & 33.54                   \\ \hline
\end{tabular}}
\caption{Inversion Ratio (IR) and Word Move Distance (WMD) of the modifications used in the experiments.}
\label{tab:shuffle-metric}
\end{table}

\begin{table*}[t]
\centering
\resizebox{0.75\textwidth}{!}{\scriptsize
\begin{tabular}{l|cc|cc|cc}
\hline
\multirow{2}{*}{source type}        & \multicolumn{2}{c|}{es} & \multicolumn{2}{c|}{ru} & \multicolumn{2}{c}{hi} \\ \cline{2-7} 
                                    & source     & target     & source     & target     & source     & target    \\ \hline
Original Language                   & 83.97      & 78.11      & 83.93      & 73.90      & 83.83      & 69.46     \\
Constituent Shuffle                 & 81.88      & 76.42      & 81.40      & 71.25      & 81.97      & 70.18     \\
Word Shuffle                        & 70.45      & 39.84      & 71.34      & 35.14      & 70.37      & 39.16     \\ 
Without Pre-training                      & 56.67      & 34.27      & 56.43      & 35.20       & 56.71      & 34.44     \\
Random Guess & 33.33 & 33.33 & 33.33 & 33.33 & 33.33 & 33.33 \\
\hline
\end{tabular}}
\caption{Comparison of XNLI results before and after removing constituent order and composition.}
\label{tab:xnli-2}
\end{table*}

\begin{table}[t]
\centering
\resizebox{0.9\columnwidth}{!}{\scriptsize
\begin{tabular}{l|c|c|c}
\hline
source type                 & es   & ru   & hi   \\ \hline
Original Language           & 82.2 & 78.1 & 60.6 \\
Constituent Shuffle         & 68.3 & 49.4 & 55.2 \\
Word Shuffle                & 37.0 & 10.6 & 15.7 \\ 
Word Embedding Average      & 32.3 & 8.5  & 13.5  \\
Random Guess                & 0.1 & 0.1 & 0.1    \\
\hline
\end{tabular}}
\caption{Comparison of Tatoeba results before and after removing constituent order and composition.}
\label{tab:Tatoeba-2}
\end{table}

\subsection{Contribution of Each Part}
We present the results in Table \ref{tab:xnli-2} and Table \ref{tab:Tatoeba-2}. we can draw the following three conclusions:\par
\textbf{On monolingual pre-training and fine-tuning, pre-training without composition still achieve good results.} We can observe that whether removing constituent order or removing composition in the source language, it still shows meaningful results (much higher than random guess) on XNLI. This illustrates that textual entailment of monolingual languages can have good performance relying only on word co-occurrence. \par
\textbf{Cross-lingual transfer works with composition and doesn't work without composition.} 
When the constituent order is removed, only a limited performance loss (within 3\%) is shown on both the source and target languages, and it is almost constant on the performance gap between source and target languages. This shows again that the contribution of constituent order to cross-language transfer is very limited and it is not a critical component of the language structure.
However, when composition is removed, the cross-lingual transfer results on target language are only slightly higher than random guess. This clearly shows that composition is the key to cross-lingual transfer. 
As for word co-occurrence, it only contributes 5\% on XNLI and 10\% to 15\% on Tatoeba. These results show that it does make some contribution but the contribution is very limited. Relying on word co-occurrence alone is not enough for a reasonable cross-lingual performance.\par

\begin{figure*}[ht]
    \centering
    \includegraphics[width=\textwidth,]{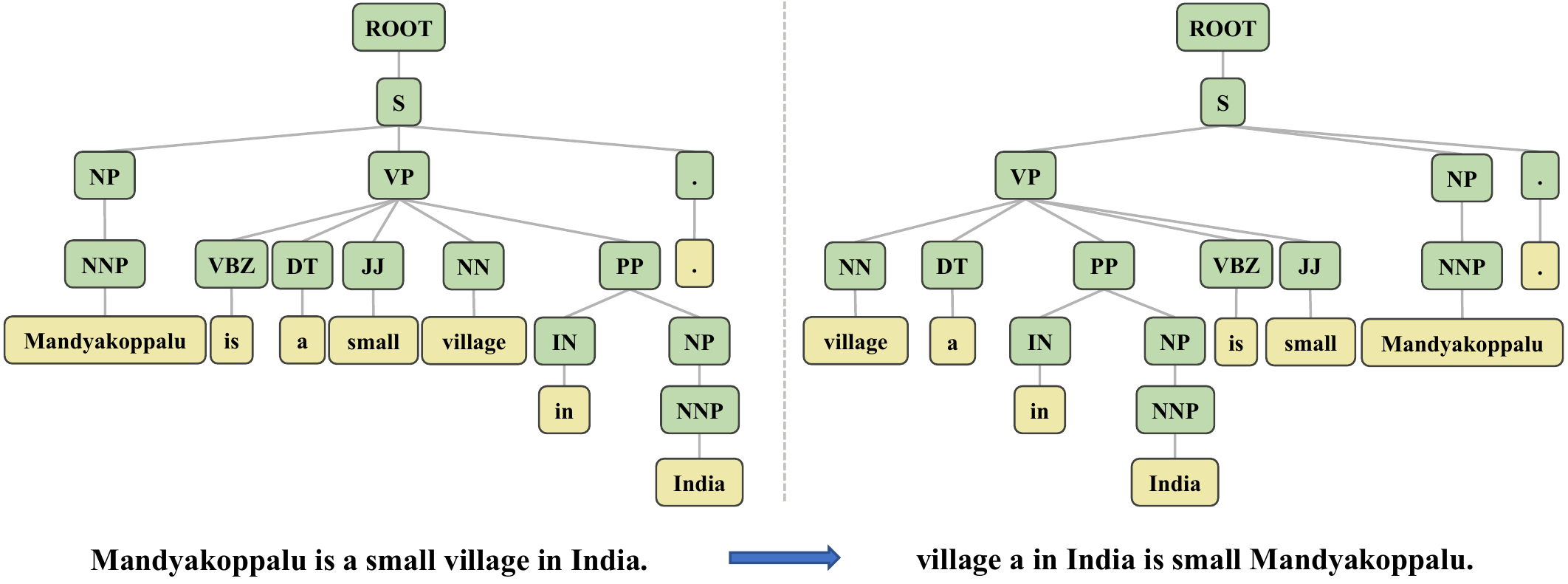}
    \caption{Example of removing intermediate nodes and disordering on the constituent tree. }
    \label{fig:example3}
\end{figure*}

\textbf{Removing constituent order and keeping composition may improve cross-lingual transfer.} We observe an interesting result in Table \ref{tab:xnli-2}. There is about 2\% drop on both Spanish and Russian after removing constituent order. However, the results show a 0.7\% improvement on Hindi while English dropped 2\%. We think this is because that model relies on every possible feature to solve English task but only relies on the commonality between language to achieve cross-lingual transfer. The model will use constituent order and composition feature to solve XNLI in unmodified English but only could use composition feature in constituent shuffled English. For languages with similar constituent order to English, more language features may lead to better performance. But for languages with different constituent order to English, only rely on composition will lead to better generalization ability. This further shows that constituent order is not key to cross-lingual transfer, and composition is the most important commonality between all languages. \par

\subsection{Detailed Analyze Composition}
To further quantify the effect of composition, we remove it in different degrees. As shown in Figure 4, we randomly remove the ratio $\alpha$ of intermediate nodes in the constituent tree. For each removed node, all its children are connected to its father. Note that an intermediate node is defined as a non-root node with more than one children.
We show Spanish results only due to space limitation. \par
\begin{table}[h]
\centering
\resizebox{\columnwidth}{!}{\scriptsize
\begin{tabular}{cc|c|c|c|c|c}
\hline
\multicolumn{2}{c|}{$\alpha$}                           & 0   & 0.25  & 0.5      & 0.75      & 1     \\ \hline
\multicolumn{1}{c|}{\multirow{2}{*}{XNLI}} & source & 82.36 & 81.44 & 81.09    & 80.66     & 70.45 \\
\multicolumn{1}{c|}{}                      & target & 76.61 & 76.19 & 75.22    & 74.17     & 39.84 \\ \hline
\multicolumn{2}{c|}{Tatoeba}                        & 83.1  & 69.8  & 62.9     & 54.8      & 47.7  \\ \hline
\end{tabular}}
\caption{Comparison of XNLI and Tatoeba results on Spanish at different ratio in removing composition similarity.}
\label{tab:rate}
\end{table}
In Table \ref{tab:rate}, We observe that when we remove 75\% of the composition, the results on XNLI are still higher than when we completely remove it. While on Tatoeba, there is a significant decrease in the results as more compostion is removed. We argue this is due to the difference in sentence length, which is much higher in XNLI than Tatoeba. Even with 75\% removed, the absolute value of retained composition is still much higher in XNLI. This result shows that only a certain ratio composition is required for reasonable performance, which shows again that composition is crucial for cross-linguistic transfer.

\section{Related Work}

\paragraph{Multilingual Pre-Training}
mBERT and XLM-R train multilingual MLM without using any parallel corpus and show strong cross-lingual ability. mBERT is an extension of BERT which is pre-trained on Wikipedia data over 100 languages to learn a language-invariant feature space shared across multiple languages. XLM-R \cite{conneau2020unsupervised} is trained on 2.5T data over 100 languages extracted from Common Crawl \cite{wenzek-etal-2020-ccnet}, which demonstrates the effect of the model trained on a large-scale corpus. Results of XLM-R on a large number of downstream cross-lingual tasks show that a large-scale training corpus can significantly improve the performance of multilingual models. \par
Other methods use parallel corpus in multilingual pre-training. XLM \cite{NEURIPS2019_cross-lingual} introduces a Translation Language Model (TLM) based on the parallel corpus which shows significant improvement on downstream tasks. Unicoder \cite{huang-etal-2019-unicoder} introduces a multitask learning framework to learn cross-lingual representations with monolingual and parallel corpora, achieving further gains. ALM \cite{yang2020alternating} allows the model to learn cross-lingual code-switch sentences, enhancing the transfer ability. Recent studies \textsc{InfoXLM} \cite{chi2021infoxlm}, \textsc{Hictl} \cite{iclr21Weion}, VECO \cite{luo2021veco} and \textsc{Ernie-M} \cite{ouyang-etal-2021-ernie} use contrastive learning, back translation and other tricks further enhancing the performance of the multilingual model.
Our work focus on studying the model only with multilingual MLM, and leave the study of works with parallel data as future work.
\paragraph{Probing Multilingual MLM}
mBERT and XLM-R have successfully achieved excellent cross-lingual transfer performance without using any parallel corpus. Researchers have wondered what the source of this cross-lingual ability is. \cite{pires2019multilingual} examines the zero-shot cross-lingual transfer performance on NER \cite{pan-etal-2017-cross} and part-of-speech (POS) tagging. They believe this success comes from the shared anchor words between languages. Not coincidentally, similar conclusion is reached by \cite{wu-dredze-2019-beto}. However, this conclusion is proven inaccurate by \cite{conneau2020emerging,Karthikeyan2020cross}. Their experiments show that the model still learns cross-lingual transfer ability on the corpus without anchor words at all. Besides, \cite{conneau2020emerging,Karthikeyan2020cross,artetxe-etal-2020-cross,libovicky-etal-2020-language,muller-etal-2021-first} have analyzed the cross-lingual ability of multilingual masked language models in terms of language similarity, shared model parameters cross languages, model structure, training objectives, language marker. The results suggest that structure similarity and shared parameters between languages are crucial for cross-lingual transfer. In this paper, we focus on analyzing language structure. We decompose it into constituent order, composition, and word co-occurrence and study the effect of each part separately.
\paragraph{Word Order in Machine Translation and Masked Language Model}
Finding the appropriate word ordering in target languages significantly influences the machine translation quality for statistical machine translation ~\cite{tillmann2004unigram, chiang2007hierarchical}, neural machine translation ~\cite{kawara2018recursive,zhao2018exploiting} and non-autoregressive neural machine translation ~\cite{ran2019guiding}. This is because the input and output sentence for machine translation have different order and the evaluation metrics also consider the output word order. While our study is different because we only focus on classification tasks and their outputs don't need to consider word order.

\citealt{ji2021word} shows that adapting word order could get about 1\% gain. While our composition reordering also could get about 1\% gain. But all these gains still show that constituent order is not important for cross-lingual transfer because that removing composition will lead to more than 30\% difference. \citealt{sinha2021masked} shows that word order is not important for English monolingual pre-training. After removing composition, our experiments also show that the performance on source languages won't drop a lot. But the performance on target language will drop more than 30\%. This proves that composition is not the key for English monolingual pre-training but the key for cross-lingual transfer.

\section{Conclusion}
In this paper, we study the source of cross-lingual ability in the multilingual masked language model in the view of language structure. 
We study three language structure properties: constituent order, composition and word co-occurrence.
The experiments are conducted using control variable method. we create an artificial language by modifying property in source language. We quantify the contribution of these three properties separately through cross-language transfer performance changes from the modified language to the target language. The results show that the contribution of constituent order and word co-occurrence are very limited, while composition is actually the key to cross-lingual transfer. How to use this finding to enhance pre-trained multilingual language models and improve performance on cross-lingual NLP tasks will be our focus for future work.

\bibliography{anthology,custom}
\bibliographystyle{acl_natbib}




\end{document}